\documentclass[11pt,a4paper]{article}
\usepackage[hyperref]{acl2018}
\usepackage{amscd}
\usepackage{amsfonts} 
\usepackage{amsmath}
\usepackage{amssymb}
\usepackage{amsthm}
\usepackage{bm}
\usepackage{booktabs}
\usepackage{breqn}
\usepackage{enumitem}
\usepackage[T1]{fontenc}
\usepackage{graphicx}
\usepackage{hyperref}
\usepackage[utf8]{inputenc}
\usepackage{latexsym}
\usepackage{microtype}
\usepackage{multirow}
\usepackage{nicefrac}
\usepackage{subcaption}
\usepackage{times}
\usepackage[normalem]{ulem}
\usepackage{url}
\usepackage{verbatim}
\usepackage{xspace}

\usepackage{tikz}
\usetikzlibrary{automata,arrows}
\usetikzlibrary{bayesnet,calc}

\newcommand{\figref}[1]{Figure~\ref{#1}}

\newif{\ifhidecomments}
\hidecommentsfalse
\ifhidecomments
    \newcommand{\nascomment}[1]{}	
    \newcommand{\dbccomment}[1]{}
    \newcommand{\chenhao}[1]{}
\else
    \newcommand{\nascomment}[1]{\textcolor{red}{[#1]}}
    \newcommand{\dbccomment}[1]{\textcolor{cyan}{[#1]}}
    \newcommand{\chenhao}[1]{\textcolor{red}{[#1 ---\textsc{ct}]}}
\fi

\newcommand{\scholar}{{\sc scholar}\xspace}

\aclfinalcopy 
\def\aclpaperid{1149} 

\title{Neural Models for Documents with Metadata}

\author{Dallas Card$^{1}$ \enskip  Chenhao Tan$^{2}$ \enskip Noah A. Smith$^{3}$ \\ 
  $^{1}$Machine Learning Department, Carnegie Mellon University, Pittsburgh, PA, 15213, USA \\
    $^{2}$Department of Computer Science, University of Colorado, Boulder, CO, 80309, USA \\
    $^{3}$Paul G. Allen School of CSE, University of Washington, Seattle, WA, 98195, USA\\
  {\tt dcard@cmu.edu} \quad {\tt chenhao.tan@colorado.edu} \\ {\tt nasmith@cs.washington.edu} \\
}

\date{}

\begin{document}
\maketitle


\begin{abstract}

Most real-world document collections involve various types of metadata, such as author, source, and date, and yet the most commonly-used approaches to modeling text corpora ignore this information. While specialized models have been developed for particular applications, few are widely used in practice, as customization typically requires derivation of a custom inference algorithm.
In this paper, 
we build on recent advances in variational inference methods and propose a general neural framework,
based on topic models,
to enable flexible incorporation of metadata and allow for rapid exploration of alternative models.
Our approach achieves strong performance, with a
manageable tradeoff between perplexity, coherence, and sparsity. 
Finally, we demonstrate the potential of our framework through an exploration of a corpus of articles about US immigration.
\end{abstract}


\section{Introduction}

Topic models comprise a family of methods for uncovering latent structure in text corpora, and are widely used tools in the digital humanities, political science, and other related fields \cite{boyd2017applications}.
Latent Dirichlet allocation (LDA; \citealp{blei.2003}) is often used when there is no prior knowledge about a corpus.
In the real world, however, most documents have non-textual attributes
such as author \cite{rosen.2004}, timestamp \cite{blei.2006}, rating \cite{mcauliffe.2008}, or ideology \cite{eisenstein.2011,nguyen.2015}, which we refer to as {\bf metadata}. 

Many customizations of LDA have been developed to incorporate document metadata.
Two models of note are \textbf{supervised LDA} (SLDA; \citealp{mcauliffe.2008}), which jointly models words and labels (e.g., ratings) as being generated from a latent representation, and \textbf{sparse additive generative} models (SAGE; \citealp{eisenstein.2011}), which assumes that observed covariates (e.g., author ideology) have a sparse effect on 
the relative probabilities of words given topics.
The structural topic model (STM; \citealp{roberts.2014}), which adds correlations between topics to SAGE, is also widely used, but like SAGE it is limited in the types of metadata it can efficiently make use of, and how that metadata is used. 
Note that in this work we will distinguish {\bf labels} (metadata that are generated jointly with words from latent topic representations)
 from  {\bf covariates} (observed metadata that influence the distribution of labels and words). 

The ability to create variations of LDA such as those listed above has 
been limited by the expertise needed to develop custom inference algorithms for each model. As a result, it is rare to see such variations being widely used in practice. 
In this work, we take advantage of recent advances in variational methods 
\cite{kingma.2014,rezende.2014,miao.2016,srivastava.2017} to facilitate approximate Bayesian inference \textit{without requiring model-specific derivations}, and
 propose a general neural framework for topic models with metadata, \scholar.\footnote{\textbf{S}parse \textbf{C}ontextual \textbf{H}idden and \textbf{O}bserved \textbf{L}anguage \textbf{A}utoencode\textbf{R}.
}

\scholar combines the abilities of SAGE and SLDA, and allows for easy exploration of the following options for customization:
\begin{enumerate}[itemsep=0pt,leftmargin=*]
\item Covariates: as in SAGE and STM, we incorporate explicit deviations for observed covariates,
as well as effects for \emph{interactions} with topics.  
\item Supervision: as in SLDA, we can use metadata as labels to help infer topics that are relevant in predicting those labels.
\item Rich encoder network: we use the encoding network of a variational autoencoder (VAE) to incorporate additional prior knowledge in the form of word embeddings, and/or to provide interpretable embeddings of covariates.
\item Sparsity: as in SAGE, a sparsity-inducing prior can be used to encourage more interpretable topics, represented as sparse deviations from a background log-frequency.
\end{enumerate}

We begin with the necessary background and motivation (\S \ref{sec:background}), and then describe our basic framework and its extensions (\S \ref{sec:model}),
followed by
a series of experiments (\S \ref{sec:experiments}).
In an unsupervised setting, we can customize the model to trade off between perplexity, coherence, and sparsity, with improved coherence through the introduction of word vectors. 
Alternatively, by incorporating metadata we can either learn topics that are more predictive of labels than SLDA, or learn explicit deviations for particular parts of the metadata.
Finally, by combining all parts of our model we can meaningfully incorporate metadata in multiple ways, which we demonstrate through an exploration of a corpus of news articles about US immigration.

In presenting this particular model, we emphasize not only its ability to adapt to the characteristics of the data, but the extent to which the VAE approach to inference provides a powerful framework for latent variable modeling that suggests the possibility of many further extensions. 
Our implementation is available at \url{https://github.com/dallascard/scholar}.


\section{Background and Motivation} \label{sec:background}

LDA can be understood as a non-negative Bayesian matrix factorization model:
the observed document-word frequency matrix, $\mathbf{X} \in \mathbb{Z}^{D \times V}$ ($D$ is the number of documents, $V$ is the vocabulary size) is factored into two low-rank matrices, $\bm{\Theta}^{D \times K}$ and $\mathbf{B}^{K \times V}$, where each row of $\bm{\Theta}$, $\bm{\theta}_i \in \Delta^K$ is a latent variable representing a distribution over topics in document $i$, and each row of $\mathbf{B}$, $\bm{\beta}_{k} \in \Delta^V$, represents a single topic, i.e., a distribution over words in the vocabulary.\footnote{$\mathbb{Z}$ denotes nonnegative integers, and $\Delta^K$ denotes the set of $K$-length nonnegative vectors that sum to one. For a proper probabilistic interpretation, the matrix to be factored is actually the matrix of latent mean parameters of the assumed data generating process, $\mathbf{X}_{ij} \sim \textrm{Poisson}(\mathbf{\Lambda}_{ij})$. See \citet{cemgil.2009} or  \citet{paisley.2014} for details.}
While it is possible to factor the count data into unconstrained matrices, 
the particular
priors
assumed by LDA
are
important for interpretability \cite{wallach.2009.nips}.
For example, the neural variational document model
(NVDM; \citealp{miao.2016}) allows $\bm{\theta}_i \in \mathbb{R}^{K}$
and achieves normalization by taking the softmax of $\bm{\theta}_i^\top \mathbf{B}$.
However, the experiments in \citet{srivastava.2017} found the performance of the NVDM to be slightly worse than LDA in terms of perplexity, and dramatically worse in terms of topic coherence.

The topics discovered by LDA tend to be parsimonious and coherent groupings of words which are readily identifiable to humans as being related to each other \cite{chang.2009}, and the resulting mode of the matrix $\bm{\Theta}$ provides a representation of each document which can be treated as a measurement for downstream tasks, such as classification or answering social scientific questions \cite{wallach.2016}. LDA does not require --- and cannot make use of --- additional prior knowledge. As such, the topics that are discovered may bear little connection to metadata of a corpus that is of interest to a researcher, such as sentiment, ideology, or time.

In this paper, we take inspiration from two models which have sought to alleviate this problem. The first, supervised LDA (SLDA; \citealp{mcauliffe.2008}), assumes that documents have labels $y$ which are generated conditional on the corresponding latent representation, i.e., ${y}_i \sim p({y} \mid \bm{\theta}_i)$.\footnote{Technically, the model conditions on the mean of the per-word latent variables, but we elide this detail in the interest of concision.} By incorporating labels into the model, it is forced to learn topics which allow documents to be represented in a way that is useful for the classification task. Such models can be used inductively as text classifiers \citep{Balasubramanyan:ProceedingsOfIcwsm:2012}.

SAGE \cite{eisenstein.2011}, by contrast, is an exponential-family model, 
where the key innovation 
was to replace topics with sparse deviations from the background log-frequency of words ($\bm{d}$),  i.e., $p(\textrm{word} \mid \textrm{softmax}(\bm{d} + \bm{\theta}_i^\top \mathbf{B}))$.
SAGE can also incorporate deviations for observed covariates, as well as interactions between topics and covariates, by including additional terms inside the softmax.
In principle, this allows for inferring, for example, the effect on an author's ideology on their choice of words, as well as ideological variations on each underlying topic.
Unlike the NVDM,
SAGE still constrains $\bm{\theta}_i$ to lie on the simplex, as in LDA.

SLDA and SAGE provide two different ways that users might wish to incorporate prior knowledge as a way of guiding the discovery of topics in a corpus:
SLDA incorporates labels through a distribution conditional on topics;
SAGE  includes explicit sparse deviations for each
unique value
 of a covariate, in addition to topics.\footnote{A third way of incorporating metadata is the approach used by various ``upstream'' models, such as Dirichlet-multinomial regression \cite{mimno.2008}, which uses observed metadata to inform the document prior. We hypothesize that this approach could be productively combined with our framework, but we leave this as future work.}

Because of the Dirichlet-multinomial conjugacy in the original model, efficient inference algorithms exist for LDA.
Each variation of LDA, however, has required the derivation of a custom inference algorithm, which is a time-consuming and error-prone process. In SLDA, for example, each type of distribution we might assume for $p({y} \mid \bm{\theta})$ would require a modification of the inference algorithm. SAGE breaks conjugacy, and as such, the authors adopted L-BFGS for optimizing the variational bound. Moreover, in order to maintain computational efficiency, it assumed that covariates were limited to a single categorical label. 

More recently, the variational autoencoder (VAE) was introduced as a way to perform approximate posterior inference on models with otherwise intractable posteriors \cite{kingma.2014,rezende.2014}. This approach has previously been applied to models of text by \citet{miao.2016} and \citet{srivastava.2017}. We build on their work and show how this framework can be adapted to seamlessly incorporate the ideas of both SAGE and SLDA, while allowing for greater flexibility in the use of metadata. Moreover, by exploiting automatic differentiation, we allow for modification of the model without requiring any change to the inference procedure.  The result is not only a highly adaptable family of models with scalable inference and efficient prediction; it also points the way to incorporation of many ideas found in the literature, such as a gradual evolution of topics \cite{blei.2006}, and hierarchical models \cite{blei.2010,nguyen.2013,nguyen.2015}.


\section{\scholar: A Neural Topic Model with Covariates, Supervision, and Sparsity} \label{sec:model}

We begin by presenting the generative story for our model, and explain how it generalizes both SLDA and SAGE (\S \ref{sec:generative}). We then provide a general explanation of inference using VAEs and how it applies to our model (\S \ref{sec:inference}), as well as how to infer document representations and predict labels at test time (\S \ref{sec:prediction}). Finally, we discuss how we can incorporate additional prior knowledge (\S \ref{sec:prior}).

\subsection{Generative Story} \label{sec:generative}
Consider  a corpus of $D$ documents, where document $i$ is a list of $N_i$ words, $\bm{w}_i$, with $V$ words in the vocabulary.
For each document, we may have observed covariates $\bm{c}_i$  (e.g., year of publication), and/or one or more labels, $\bm{y}_i$ (e.g., sentiment).

Our model builds on the generative story of LDA, but optionally incorporates labels and covariates, and replaces the matrix product of $\bm{\Theta}$ and $\mathbf{B}$ with a more flexible generative network, $f_g$, followed by a softmax transform. 
Instead of using a Dirichlet prior as in LDA,
we employ a logistic normal prior on $\bm \theta$ as in \citet{srivastava.2017} to facilitate inference (\S \ref{sec:inference}): we draw a latent variable, $\bm{r}$,\footnote{$\bm{r}$ is equivalent to $\bm{z}$ in the original VAE. To avoid confusion with topic assignment of words in the topic modeling literature, we use $\bm{r}$ instead of $\bm{z}$.} from a multivariate normal, and transform it to lie on the simplex using a softmax transform.\footnote{Unlike the correlated topic model (CTM; \citealp{lafferty.2006}), which also uses a logistic-normal prior, we fix the parameters of the prior and use a diagonal covariance matrix, rather than trying to infer correlations among topics.
However, it would be a straightforward extension of our framework to place a richer prior on the latent document representations, and learn correlations by updating the parameters of this prior after each epoch, analogously to the variational EM approach used for the CTM.} 

\begin{figure}[t]
    \centering
    \begin{subfigure}[b]{0.2\textwidth}
        \centering

\begin{tikzpicture}
  \node[obs]                               (w) {$w$};
  \node[obs, left of = w, node distance=1.2cm]   (y) {$\bm{y}$};
  \node[latent, above of = w, node distance=1.3cm, double]  (eta) {$\bm{\eta}$};
  \node[latent, above of = eta, node distance=1.3cm, double]            (theta) {$\bm{\theta}$};
  \node[latent, above of = theta, node distance=1.3cm]            (r) {$\bm{r}$};
  \node[latent, above of = r, node distance=1.2cm]          (alpha) {$\alpha$};
  \node[obs, left of = theta, yshift=-1cm, node distance=1.2cm]   (c) {$\bm{c}$};
  \node[latent, right of = theta, yshift=-1cm, node distance=1.3cm]   (beta) {$\mathbf{B}$};

  \edge {r} {theta};
  \edge {theta} {eta};
  \edge {eta} {w};
  \edge {alpha} {r};
  \edge {beta} {eta};
  \edge {c} {eta};
  \edge {c} {y};
   \edge {theta} {y};

  \plate {words} {(w)} {$N_i$} ;
  \plate {documents} {(words)(eta)(r)(y)(c)} {$D$} ;
\end{tikzpicture}
        \caption{Generative model}
        \label{fig:generative_model}
    \end{subfigure}
    \hfill
    \begin{subfigure}[b]{0.2\textwidth}
        \centering

\begin{tikzpicture}
  \node[latent, double]            (r) {$\bm{r}$};
  \node[latent, above=of r, node distance=0.6cm, xshift=-0.6cm, double]            (mu) {$\bm{\mu}$};
  \node[latent, above=of r, node distance=0.6cm, xshift=0.6cm, double]            (logsigma) {$\bm{\sigma}$};
  \node[latent, above=of mu, xshift=0.6cm, double]            (pi) {$\bm{\pi}$};
  \node[obs, above=of pi, node distance=1.0cm,  xshift=1cm]                               (c) {$\bm{c}$};
  \node[obs, above=of pi, node distance=1.0cm, xshift=-1cm]                               (y) {$\bm{y}$};
    \node[obs, above=of pi, node distance=1.0cm]                                                (w) {$\bm{w}$};
  \node[latent, below of = r, node distance=1.4cm]            (epsilon) {$\boldsymbol{\epsilon}$};

  \edge {logsigma} {r};
  \edge {mu} {r};
  \edge {w} {pi};
  \edge {y} {pi};
  \edge {c} {pi};
  \edge {epsilon} {r};
  \path[->] (pi) edge[->, >={triangle 45}] node [midway, above, sloped] {linear} (mu);
  \path[->] (pi) edge[->, >={triangle 45}] node [midway, above, sloped] {linear} (logsigma);
  \plate {documents} {(r)(mu)(logsigma)(w)(y)(c)} {$D$} ;
\end{tikzpicture}
        \caption{Inference model}
        \label{fig:vae}
    \end{subfigure}
    \caption{\figref{fig:generative_model} presents the generative story of our model.
    \figref{fig:vae} illustrates the inference network using the reparametrization trick to perform variational inference on our model. Shaded nodes are observed; double circles indicate deterministic transformations of parent nodes.  
    }
\end{figure}
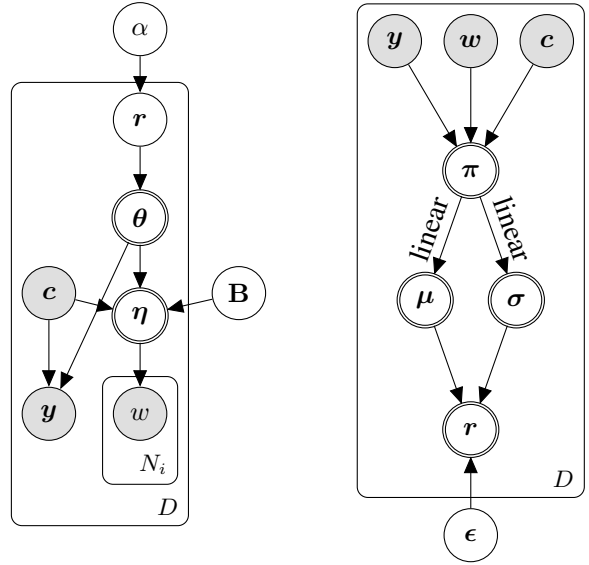

The generative story is shown in Figure \ref{fig:generative_model} and described in equations below:

\begin{itemize}[leftmargin=*,itemsep=-1pt,topsep=6pt]
\item[] For each document $i$ of length $N_i$:
\begin{itemize}[itemsep=-1pt,topsep=-3pt]
\item[] \textcolor{gray}{\# Draw a latent representation on the simplex from a logistic normal prior:}
\item[] $\bm{r}_i \sim \mathcal{N}(\bm{r} \mid \bm{\mu}_0(\alpha), \textrm{diag}(\bm{\sigma}^2_0(\alpha)))$
\item[] $\bm{\theta}_i = \textrm{softmax}(\bm{r}_i)$
\item[] \textcolor{gray}{\# Generate words, incorporating covariates:}
\item[] $\bm{\eta}_i = f_g(\bm{\theta}_i , \bm{c}_i)$
\item[] For each word $j$ in document $i$:
\begin{itemize}
\item[] $w_{ij} \sim p({w} \mid  \textrm{softmax}(\bm{\eta}_i))$
\end{itemize}
\item[] \textcolor{gray}{\# Similarly generate labels:}
\item[] $\bm{y}_i \sim p(\bm{y} \mid f_y(\bm{\theta}_i, \bm{c}_i))$,
\end{itemize}
\end{itemize}

\noindent where $p({w} \mid  \textrm{softmax}(\bm{\eta}_i))$ is a multinomial distribution and $p(\bm{y} \mid f_y(\bm{\theta}_i, \bm{c}_i))$ is a distribution appropriate to the data (e.g., multinomial for categorical labels).
$f_g$ is a model-specific combination of latent variables and covariates, $f_y$ is a multi-layer neural network, and $\bm{\mu}_0(\alpha)$ and $\bm{\sigma}_0^2(\alpha)$ are the mean and diagonal covariance terms of a multivariate normal prior.
To approximate a symmetric Dirichlet prior with hyperparameter $\alpha$, these are given by the Laplace approximation \cite{hennig.2012} to be $\mu_{0,k}(\alpha) = 0$
and
$\sigma_{0,k}^2 = (K-1)/(\alpha K)$.

If we were to ignore covariates, place a Dirichlet prior on $\mathbf{B}$, and let $\bm{\eta} = \bm{\theta}_i^\top \mathbf{B}$, this model is equivalent to SLDA with a logistic normal prior.
Similarly, we can recover a model that is like SAGE, but lacks sparsity, if we ignore labels, and let 
\begin{equation}
\bm{\eta}_i =  \bm{d} + \bm{\theta}_i^\top \mathbf{B}  + \bm{c}_i^\top \mathbf{B}^{\textit{cov}}  + (\bm{\theta}_i \otimes \bm{c}_i)^\top \mathbf{B}^{\textit{int}},
\label{eq:sage}
\end{equation}
where $\bm{d}$ is the $V$-dimensional background term (representing the log of the overall word frequency), $\bm{\theta}_i \otimes \bm{c}_i$ is a vector of interactions between topics and covariates, and $\mathbf{B}^{\textit{cov}}$ and $\mathbf{B}^{\textit{int}}$ are additional weight (deviation) matrices. The background is included to account for common words with approximately the same frequency across documents, meaning that the $\mathbf{B}^\ast$ weights now represent both positive and negative deviations from this background. This is the form of $f_g$ which we will use in our experiments.

To recover the full SAGE model, we can place a sparsity-inducing prior on each $\mathbf{B}^\ast$. As in \citet{eisenstein.2011}, we make use of the compound normal-exponential prior for each element of the weight matrices, $\mathbf{B}_{m,n}^\ast$, with hyperparameter $\gamma$,\footnote{To avoid having to tune $\gamma$, we employ an improper Jeffery's prior, $p(\tau_{m,n}) \propto 1/\tau_{m,n}$, as in SAGE. Although this causes difficulties in posterior inference for the variance terms, $\tau$, in practice, we resort to a variational EM approach, with MAP-estimation for the weights, $\mathbf{B}$, and thus alternate between computing expectations of the $\tau$ parameters, and updating all other parameters using some variant of stochastic gradient descent. For this, we only require the expectation of each $\tau_{mn}$ for each E-step, which is given by $1/\mathbf{B}_{m,n}^2$. We refer the reader to \citet{eisenstein.2011} for additional details.}
\begin{align}
\tau_{m,n} &\sim \textrm{Exponential}(\gamma), \\
\mathbf{B}_{m,n}^{\ast} &\sim \mathcal{N}(0, \tau_{m,n}).
\end{align}

We can choose to ignore various parts of this model, if, for example, we don't have any labels or observed covariates, or we don't wish to use interactions or sparsity.\footnote{We could also ignore latent topics, in which case we would get a na\"{i}ve Bayes-like model of text with deviations for each covariate $p(\bm{w}_{ij} \mid \bm{c}_i) \propto \exp (\bm{d} + \bm{c}_i^\top \mathbf{B}^{\textit{cov}})$.
}  Other generator networks could also be considered, with additional layers to represent more complex
interactions,
although this might involve some loss of interpretability. 

In the absence of metadata, and without sparsity, our model is equivalent to the ProdLDA model of \citet{srivastava.2017} with an explicit background term, and ProdLDA is, in turn, a special case of SAGE, without background log-frequencies, sparsity, covariates, or labels. In the next section we generalize the inference method used for ProdLDA; in our experiments we validate its performance and explore the effects of regularization and word-vector initialization (\S\ref{sec:prior}). The NVDM \cite{miao.2016} 
uses the same approach to inference, but 
does not not restrict document representations to the simplex.

\subsection{Learning and Inference} \label{sec:inference}

As in past work, each document $i$ is assumed to have a latent representation $\bm{r}_i$, which can be interpreted as its relative 
membership in each topic (after exponentiating and normalizing).
In order to infer an approximate posterior distribution over $\bm{r}_i$, we adopt
the sampling-based VAE
 framework developed in previous work  \citep{kingma.2014,rezende.2014}. 

As in conventional variational inference, we assume a variational approximation to the posterior, $q_{\bm \Phi}(\bm{r}_i \mid \bm{w}_i, \bm{c}_i, \bm{y}_i)$, and seek to minimize the KL divergence between it and the true posterior, $p(\bm{r}_i \mid \bm{w}_i, \bm{c}_i, \bm{y}_i)$, where $\bm \Phi$ is the set of variational parameters to be defined below. 
After some manipulations (given in supplementary materials), we obtain the evidence lower bound (ELBO) for a single document,
\begin{dmath}
{\mathcal{L}(\bm{w}_i)} = {\mathbb{E}_{q_{\bm \Phi}(\bm{r}_i\mid \bm{w}_i, \bm{c}_i, \bm{y}_i)}\left [ \sum_{j=1}^{N_i} \log p(w_{ij} \mid \bm{r}_i, \bm{c}_i) \right]} + { \mathbb{E}_{q_{\bm \Phi}(\bm{r}_i\mid \bm{w}_i, \bm{c}_i, \bm{y}_i)} \left[ \log p(\bm{y}_i \mid \bm{r}_i, \bm{c}_i) \right]} - {\textrm{D}_{\textrm{KL}} \left[q_{\bm \Phi}(\bm{r}_i \mid \bm{w}_i, \bm{c}_i, \bm{y}_i) ~||~ p(\bm{r}_i \mid \alpha) \right].}
\label{eq:bound}
\end{dmath}

As in the original VAE, we will encode the parameters of our variational distributions using a shared multi-layer neural network.
Because we have assumed a diagonal normal prior on $\bm{r}$, this will take the form of a network which outputs a mean vector, $\bm{\mu}_i = f_{\bm\mu}(\bm{w}_i, \bm{c}_i, \bm{y}_i)$ and diagonal of a covariance matrix, $\bm{\sigma}^2_i = f_{\bm\sigma}(\bm{w}_i, \bm{c}_i, \bm{y}_i)$, such that $q_{\bm\Phi}(\bm{r}_i \mid \bm{w}_i, \bm{c}_i, \bm{y}_i) = \mathcal{N}(\bm{\mu}_i, \bm{\sigma}_i^2)$.
Incorporating labels and covariates to the inference network used by \citet{miao.2016} and \citet{srivastava.2017}, we use:
\begin{align}
\bm{\pi}_i &= f_e([\mathbf{W}_x \bm{x}_i ; \mathbf{W}_c \bm{c}_i; \mathbf{W}_y \bm{y}_i]) \label{eq:pi}, \\
\bm{\mu}_i &= \mathbf{W}_{\mu} \bm{\pi}_i + \bm{b}_\mu, \\
\log \bm{\sigma}_i^2 &= \mathbf{W}_{\sigma} \bm{\pi}_i + \bm{b}_{\sigma}, \label{eq:sigma}
\end{align}
where $\bm{x}_i$ is a $V$-dimensional vector representing the counts of words in $\bm{w}_i$, and $f_e$ is a multilayer perceptron. The full set of encoder parameters, $\bm{\Phi}$, thus includes the parameters of $f_e$ and all weight matrices and bias vectors in Equations~\ref{eq:pi}--\ref{eq:sigma} (see Figure~\ref{fig:vae}).

This approach means that the expectations in Equation~\ref{eq:bound} are intractable, but we can approximate them using sampling. In order to maintain differentiability with respect to $\bm{\Phi}$, even after sampling, we make use of the reparameterization trick \citep{kingma.2014},\footnote{
The Dirichlet distribution cannot be directly reparameterized in this way, 
which 
is why we use the logistic normal prior on $\bm{\theta}$
to approximate the
Dirichlet prior used in LDA.}  which allows us to reparameterize samples from $q_{\bm\Phi}(\bm{r} \mid \bm{w}_i, \bm{c}_i, \bm{y}_i)$ in terms of samples from an independent source of noise, i.e.,
\begin{align*}
\bm{\epsilon}^{(s)} &\sim \mathcal{N}(0, \mathbf{I}),\\
\bm{r}_i^{(s)} = g_{\bm{\Phi}}(\bm{w}_i, \bm{c}_i, \bm{y}_i, \bm{\epsilon}^{(s)}) &= \bm{\mu}_i + \bm{\sigma}_i \cdot \bm{\epsilon}^{(s)}.
\end{align*}

We thus replace the bound in Equation~\ref{eq:bound} with a Monte Carlo approximation using a single sample\footnote{Alternatively, one can average over multiple samples.} of $\bm{\epsilon}$ (and thereby of $\bm{r}$):
\begin{dmath}
\mathcal{L}(\bm{w}_i) \approx {\sum_{j=1}^{N_i} \log p(w_{ij} \mid \bm{r}_i^{(s)}, \bm{c}_i)  +   \log p(\bm{y}_i \mid \bm{r}_i^{(s)}, \bm{c}_i)} - {\textrm{D}_{\textrm{KL}} \left[q_{\bm{\Phi}}(\bm{r}_i \mid \bm{w}_i, \bm{c}_i, \bm{y}_i)~||~p(\bm{r}_i \mid \alpha) \right].}
\label{eq:approx_bound}
\end{dmath}
We can now optimize this sampling-based approximation of the variational bound with respect to $\bm{\Phi}$, $\mathbf{B}^\ast$, and all parameters of $f_g$ and $f_y$ using stochastic gradient descent.
Moreover, because of this stochastic approach to inference, we are \textit{not} restricted to covariates with a small number of unique values, which was a limitation of SAGE.
Finally, the KL divergence term in Equation \ref{eq:approx_bound} can be computed in closed form (see supplementary materials).

\subsection{Prediction on Held-out Data} \label{sec:prediction}
In addition to inferring latent topics, our model can both infer latent representations for new documents and predict their labels, the latter of which was the motivation for SLDA. In traditional variational inference, inference at test time requires fixing global parameters (topics), and optimizing the per-document variational parameters for the test set. 
With the VAE framework, 
by contrast,
the encoder network
(Equations~\ref{eq:pi}--\ref{eq:sigma})
can 
be used
 to directly estimate the posterior distribution 
for each
 test document,
  using only a forward pass (no iterative optimization or sampling).

If not using labels, we can use this approach directly, passing the word counts of new documents through the encoder to get a posterior $q_{\bm\Phi}(\bm{r}_i \mid \bm{w}_i, \bm{c}_i)$. When we also include labels to be predicted, we can first train a fully-observed model, as above, then fix the decoder, and retrain the encoder without labels.
In practice, however, if we train the encoder network using  one-hot encodings of document labels, it is sufficient to provide a vector of all zeros for the labels of test documents; this is what we 
adopt for
our experiments (\S \ref{sec:classification}), and we still obtain good predictive performance.

The label network, $f_y$, is a flexible component which can be used to predict a wide range of outcomes, from categorical labels (such as star ratings; \citealp{mcauliffe.2008}) 
to real-valued outputs (such as number of citations or box-office returns; \citealp{yogatama2011predicting}).
For categorical labels, predictions are given by
\begin{align}
\hat y_i = \underset{y~ \in~ \mathcal{Y}}{\textrm{argmax}}~ p (y \mid \bm{r}_i, \bm{c}_i). \label{eq:joint}
\end{align}

Alternatively, when dealing with a small set of categorical labels, it is also possible to treat them as observed categorical covariates during training. At test time, we can then consider all possible one-hot vectors, $\bm{e}$, in place of $\bm{c}_i$, and predict the label that maximizes the probability of the words, i.e.,
\begin{align}
\hat y_i = \underset{y ~\in~ \mathcal{Y}}{\textrm{argmax}}~ \sum_{j=1}^{N_i} \log p(w_{ij} \mid \bm{r}_i, \bm{e}_y). \label{eq:conditional}
\end{align}
This approach works well in practice (as we show in \S \ref{sec:classification}),
but does not scale to large numbers of labels, or other types of prediction problems, such as multi-class classification or regression. 

The choice to include metadata as covariates, labels, or both, depends on the data.
The key point is that we can incorporate metadata in two very different ways, depending on what we want from the model. Labels guide the model to infer topics that are relevant to those labels, whereas covariates induce explicit deviations, leaving the latent variables to account for the rest of the content.

\subsection{Additional Prior Information} \label{sec:prior}
A final advantage of the VAE framework is that the encoder network provides a way to incorporate additional prior information in the form of word vectors. Although we can learn all parameters starting from a random initialization, it is also possible to initialize and fix the initial embeddings of words in the model, $\mathbf{W}_x$, in Equation~\ref{eq:pi}. This leverages word similarities derived from large amounts of unlabeled data, and may promote greater coherence in inferred topics. The same could also be done for some covariates; for example, we could embed the source of a news article based on its place on the ideological spectrum. Conversely, if we choose to learn these parameters, the learned values ($\mathbf{W}_y$ and $\mathbf{W}_c$) may provide meaningful embeddings of these metadata (see section \S \ref{sec:framing}).

Other variants on topic models have also proposed incorporating word vectors, both as a parallel part of the generative process \cite{nguyen.2015.acl}, and as an alternative parameterization of topic distributions \cite{das.2015}, but inference is not scalable in either of these models. Because of the generality of the VAE framework, we could also modify the generative story so that word embeddings are emitted (rather than tokens);
we leave this for future work.


\section{Experiments and Results} \label{sec:experiments}

To evaluate and demonstrate the potential of this model, we present a series of experiments below. We first test \scholar without observed metadata, and explore the effects of using regularization and/or word vector initialization, compared to LDA, SAGE, and NVDM (\S\ref{sec:unsup}). We then evaluate our model in terms of predictive performance, in comparison to SLDA and an $l_2$-regularized logistic regression baseline (\S\ref{sec:classification}). Finally, we demonstrate the ability to incorporate covariates and/or labels in an exploratory data analysis (\S\ref{sec:framing}). 

The scores we report are generalization to held-out data, measured in terms of perplexity; coherence, measured in terms of non-negative point-wise mutual information (NPMI; \citealp{chang.2009,newman.2010}), and classification accuracy on test data. For coherence we evaluate NPMI using the top $10$ words of each topic, both internally (using test data), and externally, using a decade of articles from the English Gigaword dataset \citep{graff2003english}. 
Since our model employs variational methods, the reported perplexity is an upper bound based on the ELBO.

As datasets we use the familiar 20 newsgroups, the IMDB corpus of 50,000 movie reviews \cite{maas.2011}, and the UIUC Yahoo answers dataset with 150,000 documents in 15 categories \cite{chang.2008}.
For further exploration, we also make use of a corpus of approximately 4,000 time-stamped news articles about US immigration, each annotated with pro- or anti-immigration tone \cite{card.2015}. We use the original author-provided implementations of SAGE\footnote{\url{github.com/jacobeisenstein/SAGE}} and SLDA,\footnote{\url{github.com/blei-lab/class-slda}} while for LDA we use Mallet.\footnote{\url{mallet.cs.umass.edu}}.
Our implementation of \scholar is in TensorFlow, but we have also provided a preliminary PyTorch implementation of the core of our model.\footnote{\url{github.com/dallascard/scholar}}
For additional details about datasets and implementation, please refer to the supplementary material.

It is challenging to fairly evaluate the relative computational efficiency of our approach compared to past work (due to the stochastic nature of our approach to inference, choices about hyperparameters such as tolerance, and because of differences in implementation). Nevertheless, in practice, the performance of our approach is highly appealing. For all experiments in this paper, our implementation was much faster than SLDA or SAGE (implemented in C and Matlab, respectively), and competitive with Mallet.

\subsection{Unsupervised Evaluation} \label{sec:unsup}

Although the emphasis of this work is on incorporating observed labels and/or covariates, we briefly report on experiments in the unsupervised setting.
Recall that, without metadata, \scholar equates to ProdLDA, but with an explicit background term.\footnote{Note, however, that a batchnorm layer in ProdLDA may play a similar role to a background term, and there are small differences in implementation; please see supplementary material for more discussion of this.} 
We therefore use the same experimental setup as \citet{srivastava.2017} (learning rate, momentum, batch size, and number of epochs) and find the same general patterns as they reported (see Table \ref{tab:unsupervised} and supplementary material): our model returns more coherent topics than LDA, but at the cost of worse perplexity. SAGE, by contrast, attains very high levels of sparsity, but at the cost of worse perplexity and coherence than LDA. As expected, the NVDM produces relatively low perplexity, but very poor coherence, due to its lack of constraints on $\bm{\theta}$.

Further experimentation revealed that the VAE framework involves a tradeoff among the scores; running for more epochs tends to result in better perplexity on held-out data, but at the cost of worse coherence. Adding regularization to encourage sparse topics has a similar effect as in SAGE, leading to worse perplexity and coherence, but it does create sparse topics. Interestingly, initializing the encoder with pretrained word2vec embeddings, and \textit{not} updating them returned a model with the best
internal
coherence of any model we considered for IMDB and Yahoo answers, 
and the second-best for 20 newsgroups.

The background term in our model does not have much effect on perplexity, but plays an important role in producing coherent topics; as in SAGE, the background can account for common words, so they are mostly absent among the most heavily weighted words in the topics.
For instance, words like \textit{film} and \textit{movie} in the IMDB corpus are relatively unimportant in the topics learned by our model, but would be much more heavily weighted without the background term, as they are in topics learned by LDA.

\begin{table}
\small
\centering
\begin{tabular}{l  c  c  c  c}
 & Ppl. & NPMI  & NPMI &  Sparsity \\
Model & $\downarrow$ & (int.) $\uparrow$  & (ext.) $\uparrow$ & $\uparrow$ \\
\midrule
LDA & \bf1508 & 0.13 & 0.14  & 0 \\
SAGE & 1767 & 0.12 &  0.12 & \bf0.79 \\
NVDM & 1748 & 0.06 & 0.04 & 0 \\
\scholar $-$ \sc b.g. & 1889 & 0.09 & 0.13 & 0 \\
\scholar & 1905 &  0.14 & 0.13 & 0 \\
\scholar + \sc w.v. & 1991 & \bf0.18 & \bf0.17  & 0 \\
\scholar + \sc reg. & 2185 & 0.10 & 0.12 & 0.58 \\
\end{tabular}
\caption{Performance of our various models in an unsupervised setting (i.e., without labels or covariates) on the IMDB dataset using a 5,000-word vocabulary and 50 topics. The supplementary materials contain additional results for 20 newsgroups and Yahoo answers.
}
\label{tab:unsupervised}
\end{table}

\subsection{Text Classification} \label{sec:classification}

We next consider the utility of our model in the context of categorical labels, and consider them alternately as observed covariates and as labels generated conditional on the latent representation. We use the same setup as above, but tune number of training epochs for our model using a random 20\% of training data as a development set, and similarly tune regularization for logistic regression.

Table \ref{tab:labels} summarizes the accuracy of various models on three datasets, revealing that our model offers competitive performance, both as a joint model of words and labels (Eq.~\ref{eq:joint}), and a model which conditions on covariates (Eq.~\ref{eq:conditional}). Although \scholar is comparable to the logistic regression baseline, our purpose here is not to attain state-of-the-art performance on text classification.
Rather, the high accuracies we obtain demonstrate that we are learning low-dimensional representations of documents that are relevant to the label of interest, outperforming SLDA, and have the same attractive properties as topic models. Further, any neural network that is successful for text classification could be incorporated into $f_y$ and trained end-to-end along with topic discovery.

\begin{table}
\small
\centering
\begin{tabular}{l c c c }
& 20news & IMDB & Yahoo \\
Vocabulary size & 2000 & 5000 & 5000 \\
Number of topics & 50 & 50 & 250 \\
\midrule
SLDA & 0.60 & 0.64 & 0.65 \\
\scholar (labels) & 0.67 & 0.86 & 0.73 \\
\scholar (covariates) & \bf0.71 & \bf0.87 & 0.72 \\
Logistic regression &  0.70  & \bf0.87 & \bf0.76 \\
\end{tabular}
\caption{Accuracy of various models on three datasets with categorical labels.
}
\label{tab:labels}
\end{table}

\subsection{Exploratory Study}
\label{sec:framing}

We demonstrate how our model might be used to explore an annotated corpus of articles about immigration, and adapt to different assumptions about the data. We only use a small number of topics in this part ($K=8$) for compact presentation.

\begin{figure}[t]
\includegraphics[scale=0.56]{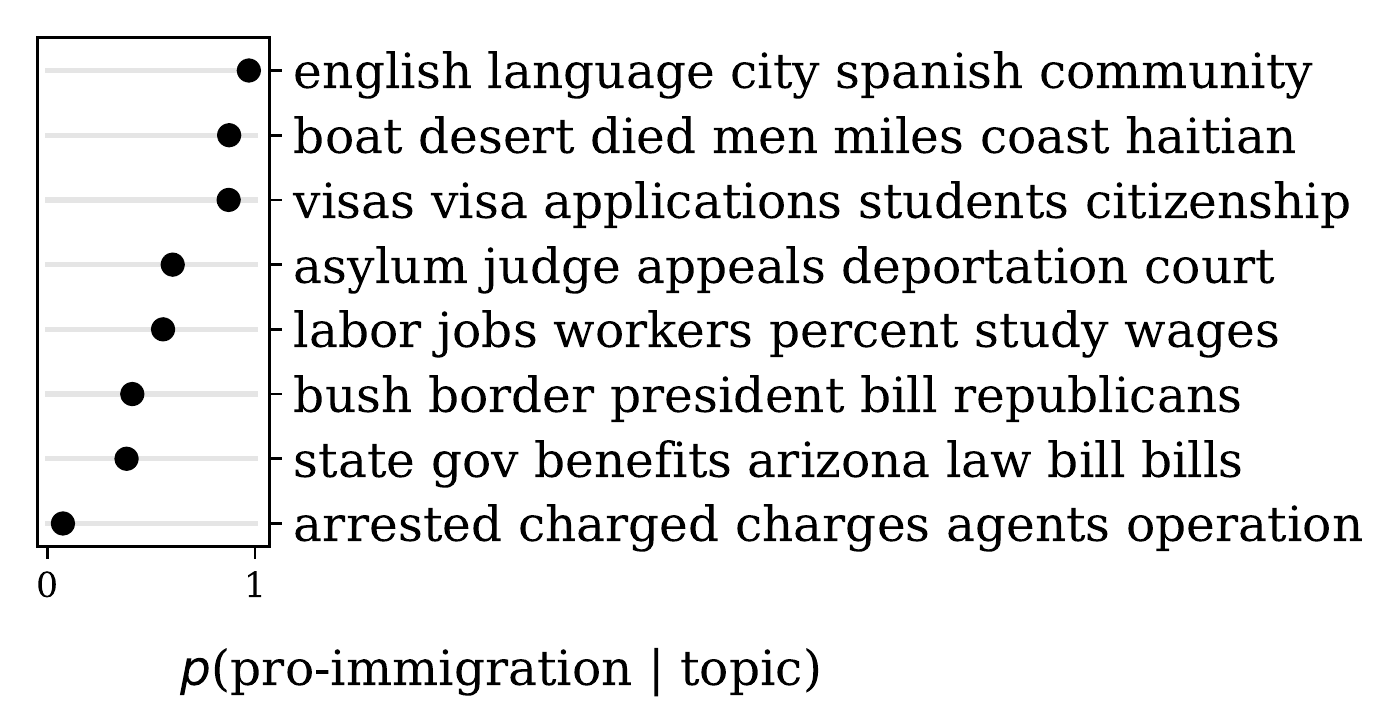}
\caption{Topics inferred by a joint model of words and tone, and the corresponding probability of 
pro-immigration tone for each topic. A topic is represented by the top words sorted by word probability throughout the paper.
}
\label{fig:tone}
\end{figure}

\paragraph{Tone as a label.}
We first consider using the annotations as a \emph{label}, and train a joint model to infer topics relevant to the tone of the article (pro- or anti-immigration). Figure \ref{fig:tone} shows a set of topics learned in this way, along with the predicted probability of an article being pro-immigration conditioned on the given topic.
All topics are coherent, and the predicted probabilities have strong face validity, e.g., ``arrested charged charges agents operation'' is least associated with pro-immigration.

\begin{table*}
\small
\centering
\begin{tabular}{l l l }
Base topics (each row is a topic) & Anti-immigration interactions & Pro-immigration interactions \\
\hline
ice customs agency enforcement homeland &  {\em \bf criminal} customs arrested  & {\em \bf detainees} detention center agency \\
population born percent americans english & jobs million {\em \bf illegals} taxpayers & english {\em \bf newcomers} hispanic city \\
judge case court guilty appeals attorney  & guilty charges man charged  & asylum court judge case appeals \\
patrol border miles coast desert boat guard & patrol border agents boat   & died authorities desert border bodies \\
licenses drivers card visa cards applicants  & foreign sept visas system & green citizenship card citizen apply  \\
island story chinese ellis international  & smuggling federal charges  & island school ellis english story  \\
guest worker workers bush labor bill & bill border house senate  & workers tech skilled farm labor  \\
benefits bill welfare republican state senate  & republican california gov state & law welfare students tuition 
\end{tabular}
\vspace{-1em}
\caption{Top words for topics (left) and the corresponding anti-immigration (middle) and pro-immigration (right) variations when treating tone as a covariate, with interactions.}
\label{tab:inter}
\end{table*}

\paragraph{Tone as a covariate.}
Next we consider using tone as a \emph{covariate}, and build a model using both tone and tone-topic interactions. Table \ref{tab:inter} shows a set of topics learned from the immigration data, along with the most highly-weighted words in the corresponding tone-topic interaction terms.
As can be seen, these interaction terms tend to capture different frames (e.g., ``criminal'' vs.~``detainees'',  and ``illegals'' vs. ``newcomers'', etc).

\begin{figure}[t]
\includegraphics[scale=0.65]{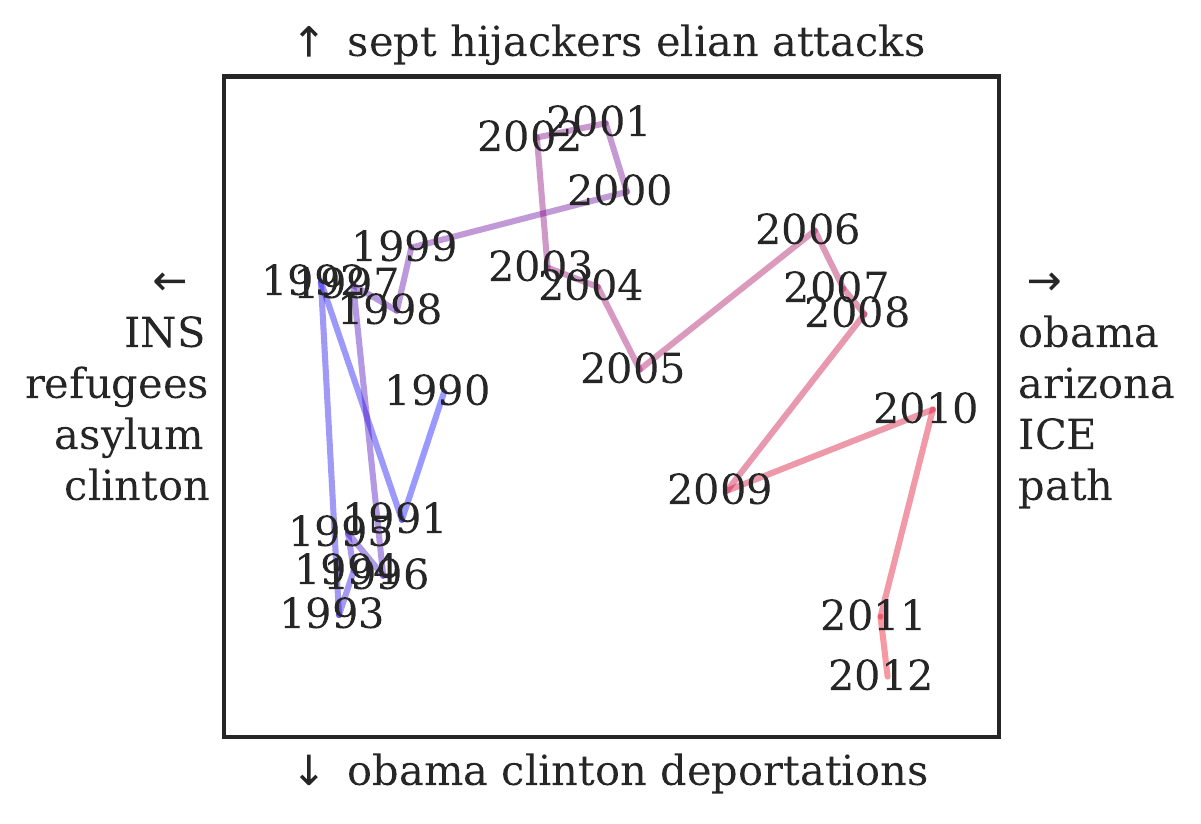}
\caption{Learned embeddings of year-of-publication (treated as a covariate) from combined model of news articles about immigration.
}
\label{fig:time}
\end{figure}

\paragraph{Combined model with temporal metadata.}
Finally, 
we incorporate both the tone annotations and the year of publication of each article, treating the former as a label and the latter as a covariate.
In this model, we also include an embedding matrix, $\mathbf{W}_c$, to project the one-hot {\em year} vectors down to a two-dimensional continuous space, with a learned deviation for each dimension.
We omit the topics in the interest of space, but 
Figure \ref{fig:time} shows the learned embedding for each year, along with the top terms of the corresponding deviations. As can be seen, the model learns that adjacent years tend to produce similar deviations, {\it even though we have not explicitly encoded this information}. The left-right dimension roughly tracks a temporal trend with positive deviations shifting from the years of \emph{Clinton} and \emph{INS} on the left, to \emph{Obama} and \emph{ICE} on the right.\footnote{The Immigration and Naturalization Service (INS) was transformed into Immigration and Customs Enforcement (ICE) and other agencies in 2003.}  Meanwhile, the events  of 9/11 dominate the vertical direction, with the words \emph{sept}, \emph{hijackers}, and \emph{attacks} increasing in probability as we move up in the space. If we wanted to look at each year individually, we could drop the embedding of years, and learn a sparse set of topic-year interactions, 
similar to tone in Table \ref{tab:inter}.

\section{Additional Related Work}

The literature on topic models is vast; in addition to papers cited throughout, other efforts to incorporate metadata into topic models include Dirichlet-multinomial regression (DMR; \citealp{mimno.2008}), Labeled LDA \cite{ramage.2009}, and MedLDA \cite{zhu.2009}. A recent paper also extended DMR by using deep neural networks to embed metadata into a richer document prior \cite{benton.2018}. 

A separate line of work has pursued parameterizing unsupervised models of documents using neural networks \cite{hinton.2009,larochelle.2012}, including non-Bayesian approaches \cite{cao.2015}. More recently, \citet{lau.2017} proposed a neural language model that incorporated topics, and \citet{he.2017} developed a scalable alternative to the correlated topic model by simultaneously learning topic embeddings.

Others have attempted to extend the reparameterization trick to the Dirichlet and Gamma distributions, either through transformations  \citep{kucukelbir.2016} or a generalization of reparameterization \citep{ruiz.2016}. Black-box and VAE-style inference have been implemented in at least two general purpose tools designed to allow rapid exploration and evaluation of models \citep{kucukelbir.2015,tran.2016}.


\section{Conclusion}

We have presented a neural framework for generalized topic models to enable 
flexible incorporation of metadata with a variety of options.
We take advantage of stochastic variational inference to develop a general algorithm for our framework such that variations do not require any model-specific algorithm derivations.
Our model demonstrates the tradeoff between perplexity, coherence, and sparsity, and outperforms SLDA in predicting document labels.
Furthermore, the flexibility of our model enables intriguing exploration of a text corpus on US immigration.
We believe that our model and code will facilitate rapid exploration of document collections with metadata.

\section*{Acknowledgments}

We would like to thank Charles Sutton, anonymous reviewers, and all members of Noah's ARK for helpful discussions and feedback. This work was made possible by a University of Washington Innovation award and computing resources provided by XSEDE.

\bibliographystyle{acl_natbib}
\bibliography{refs}


\section*{Supplementary Material}

\renewcommand{\thesubsection}{\Alph{subsection}}

\subsection{Deriving the ELBO}

The derivation of the ELBO for our model is given below, dropping explicit reference to $\bm{\Phi}$ and $\alpha$ for simplicity. For document $i$,
\begin{align}
&\log~p(\bm{w}_i, \bm{y}_i \mid \bm{c}_i) =  \log \int_{\bm{r}_i}  p(\bm{w}_{i}, \bm{y}_i, \bm{r}_i \mid \bm{c}_i) d\bm{r}_i \\
&=  \log \int_{\bm{r}_i}  p(\bm{w}_{i}, \bm{y}_i, \bm{r}_i \mid, \bm{c}_i) \frac{q(\bm{r}_i \mid \bm{w}_i, \bm{c}_i, \bm{y}_i)}{q(\bm{r}_i \mid \bm{w}_i, \bm{c}_i, \bm{y}_i)}  d\bm{r}_i\\
&=  \log \left( \mathbb{E}_{q(\bm{r}_i \mid \bm{w}_i, \bm{c}_i, \bm{y}_i)} \left[ \frac{p(\bm{w}_{i}, \bm{y}_i, \bm{r}_i \mid \bm{c}_i)}{q(\bm{r}_i \mid \bm{w}_i, \bm{c}_i, \bm{y}_i)} \right] \right)\\
&\begin{aligned}
\geq \mathbb{E}_{q(\bm{r}_i \mid \bm{w}_i, \bm{c}_i, \bm{y}_i)} \left[\log p(\bm{w}_{i}, \bm{y}_i, \bm{r}_i \mid \bm{c}_i)\right]
\\ - \mathbb{E}_{q(\bm{r}_i \mid \bm{w}_i, \bm{c}_i, \bm{y}_i)} \left[\log q(\bm{r}_i \mid \bm{w}_i, \bm{c}_i, \bm{y}_i) \right]
\end{aligned}\\
&\begin{aligned}
= \mathbb{E}_{q(\bm{r}_i \mid \bm{w}_i, \bm{c}_i, \bm{y}_i)} &\left[\log p(\bm{w}_{i}, \bm{y}_i \mid  \bm{r}_i, \bm{c}_i)\right]
\\ + \mathbb{E}_{q(\bm{r}_i \mid \bm{w}_i, \bm{c}_i, \bm{y}_i)} &\left[\log p(\bm{r}_i)\right]
\\  - \mathbb{E}_{q(\bm{r}_i \mid \bm{w}_i, \bm{c}_i, \bm{y}_i)} &\left[\log q(\bm{r}_i \mid \bm{w}_i, \bm{c}_i, \bm{y}_i) \right]
\end{aligned}\\
&\begin{aligned}
= \mathbb{E}_{q(\bm{r}_i \mid \bm{w}_i, \bm{c}_i, \bm{y}_i)} &\left[ \sum_{j=1}^{N_i} \log p(w_{ij} \mid  \bm{r}_i, \bm{c}_i)\right]
\\ + \mathbb{E}_{q(\bm{r}_i \mid \bm{w}_i, \bm{c}_i, \bm{y}_i)} &\left[\log p(\bm{y}_i \mid  \bm{r}_i, \bm{c}_i)\right]
\\ - \textrm{D}_{\textrm{KL}} &\left[ q(\bm{r}_i \mid \bm{w}_i, \bm{c}_i, \bm{y}_i) ~||~ p(\bm{r}_i ) \right] 
\end{aligned}
\end{align}

\subsection{Model details}

The KL divergence term in the variational bound can be computed as 
\begin{dmath} 
\textrm{D}_{\textrm{KL}}[q_{\bm\Phi}(\bm{r}_i \mid \bm{w}_i, \bm{c}_i, \bm{y}_i) \| p(\bm{r}_i)] = \frac{1}{2} \left(\textrm{tr}(\bm{\Sigma}_0^{-1} \bm\Sigma_i) + (\bm\mu_i - \bm\mu_0)^\top \bm\Sigma_0^{-1} (\bm\mu_i - \bm\mu_0) - K + \log \frac{|\bm\Sigma_0|}{| \bm\Sigma_i |} \right)
\end{dmath}
where $\bm\Sigma_i = \textrm{diag}(\bm\sigma_i^2(\bm{w}_i, \bm{c}_i, \bm{y}_i))$, and $\bm\Sigma_0 = \textrm{diag}(\bm\sigma_0^2(\alpha))$.

\subsection{Practicalities and Implementation} \label{sec:implementation}
As observed in past work, inference using a VAE can suffer from component collapse, which translates into excessive redundancy in topics (i.e., many topics containing the same set of words). To mitigate this problem, we borrow the approach used by \citet{srivastava.2017}, and make use of the Adam optimizer with a high momentum, combined with batchnorm layers to avoid divergence. Specifically, we add batchnorm layers following the computation of $\bm{\mu}$, $\log \bm{\sigma}^2$, and $\bm{\eta}$.

This effectively solves the problem of mode collapse, but the batchnorm layer on $\bm{\eta}$ introduces an additional problem, not previously reported. At test time, the batchnorm layer will shift the input based on the learned population mean of the training data; this effectively encodes information about the distribution of words in this model that is not captured by the topic weights and background distribution. As such, although reconstruction error will be low, the document representation $\bm{\theta}$, will not necessarily be a useful representation of the topical content of each document. In order to alleviate this problem, we reconstruct $\bm{\eta}$ as a convex combination of two copies of the output of the generator network, one passed through a batchnorm layer, and one not. During training, we then gradually anneal the model from relying entirely on the component passed through the batchnorm layer, to relying entirely on the one that is not. This ensures that the the final weights and document representations will be properly interpretable.

Note that although ProdLDA \cite{srivastava.2017} does not explicitly include a background term, it is possible that the batchnorm layer applied to $\bm{\eta}$ has a similar effect, albeit one that is not as easily interpretable. This annealing process avoids that ambiguity. 

\subsection{Data} 

All datasets were preprocessed by tokenizing, converting to lower case, removing punctuation, and dropping all tokens that included numbers, all tokens less than 3 characters, and all words on the stopword list from the snowball sampler.\footnote{\url{snowball.tartarus.org/algorithms/english/stop.txt}} The vocabulary was then formed by keeping the words that occurred in the most documents (including train and test), up to the desired size (2000 for 20 newsgroups, 5000 for the others). Note that these small preprocessing decisions can make a large difference to perplexity. We therefore include our preprocessing scripts as part of our implementation so as to facilitate easy future comparison. 

For the UIUC Yahoo answers dataset, we downloaded the documents from the project webpage.\footnote{\url{cogcomp.org/page/resource_view/89}} However, the file that is available does not completely match the description on the website. We dropped  \textit{Cars and Transportation} and \textit{Social Science} which had less than the expected number of documents, and merged \textit{Arts} and \textit{Arts and Humanities}, which appeared to be the same category, producing 15 categories, each with 10,000 documents.

\subsection{Experimental Details}

For all experiments we use a model with 300-dimensional embeddings of words, and we take $f_e$ to be only the element-wise softplus nonlinearity (followed by the linear transformations for $\bm{\mu}$ and $\log \bm{\sigma}^2$). Similarly, $f_y$ is a linear transformation of $\bm \theta$, followed by a softplus layer, followed by a linear transformation to the size of the output (the number of classes). During training, we set $S$ (the number of samples of  $\bm\epsilon$) to be 1; for estimating the ELBO at on test documents, we set $S=20$.

For the unsupervised results, we use the same set up as \cite{srivastava.2017}: Adam optimizer with $\beta_1=0.99$, learning rate $=0.002$, batch size of $200$, and training for $200$ epochs. The setup was the same for all datasets, except we only trained for $150$ epochs on Yahoo answers because it is much larger.  For LDA, we updated the hyperparameters every 10 epochs.

For the external evaluation of NPMI, we use the co-occurrence statistics from all New York Times articles in the English Gigaword published from the start of 2000 to the end of 2009, processed in the same way as our data.

For the text classification experiments, we use the \texttt{scikit-learn} implementation of logistic regression. We give it access to the same input data as our model (using the same vocabulary), and tune the strength of $l_2$ regularization using cross-validation. For our model, we only tune the number of epochs, evaluating on development data. Our models for this task did not use regularization or word vectors. 

\subsection{Additional Experimental Results}

In this section we include additional experimental results in the unsupervised setting.

Table \ref{tab:20news} shows results on the 20 newsgroups dataset, using 20 topics with a 2,000-word vocabulary. Note that these results are not necessarily directly comparable to previously published results, as preprocessing decisions can have a large impact on perplexity. These results show a similar pattern to those on the IMDB data provided in the main paper, except that word vectors do not improve internal coherence on this dataset, perhaps because of the presence of relatively more names and specialized terminology. Also, although the NVDM still has worse perplexity than LDA, the effects are not as dramatic as reported in \cite{srivastava.2017}. Regularization is also more beneficial for this data, with both SAGE and our regularized model obtaining better coherence than LDA. The topics from \textsc{scholar} for this dataset are also shown in Table \ref{tab:topics}.
 
\begin{table}[t]
\small
\centering
\begin{tabular}{l  c  c  c  c}
 & Ppl. & NPMI  & NPMI &  Sparsity \\
Model  & $\downarrow$ & (int.) $\uparrow$  & (ext.) $\uparrow$ & $\uparrow$ \\
\hline 
LDA &  \bf810 & 0.20 & 0.11 & 0 \\
SAGE & 867 & 0.27 & 0.15 & \bf0.71 \\ 
NVDM & 1067 & 0.18 & 0.11 & 0 \\
\sc scholar - b.g. & 928 & 0.17 & 0.09 & 0 \\
\sc scholar & 921 & \bf0.35 & 0.16 & 0 \\
\sc scholar + w.v. & 955 & 0.29 & \bf0.17 & 0 \\
\sc scholar + reg. & 1053 & 0.25 & 0.13 & 0.43 \\
\end{tabular}
\caption{Performance of various models on the 20 newsgroups dataset with 20 topics and a 2,000-word vocabulary.}
\label{tab:20news}
\end{table}

Table \ref{tab:yahoo} shows the equivalent results for the Yahoo answers dataset, using 250 topics, and a 5,000-word vocabulary. These results closely match those for the IMDB dataset, with our model having higher perplexity but also higher internal coherence than LDA. As with IMDB, the use of word vectors improves coherence, both internally and externally, but again at the cost of worse perplexity. Surprisingly, our model without a background term actually has the best \textit{external} coherence on this dataset, but as described in the main paper, these tend to give high weight primarily to common words, and are more repetitive as a result.

\begin{table}[t]
\small
\centering
\begin{tabular}{l  c  c  c  c}
 & Ppl. & NPMI  & NPMI &  Sparsity \\
Model  & $\downarrow$ & (int.) $\uparrow$  & (ext.) $\uparrow$ & $\uparrow$ \\
\hline 
LDA &  \bf1035 & 0.29 & 0.15 & 0 \\
NVDM & 4588 & 0.20 & 0.09 & 0 \\
\sc scholar - b.g. & 1589 & 0.27 & \bf0.16 & 0 \\
\sc scholar & 1596 & 0.33 & 0.13 & 0 \\
\sc scholar + w.v. & 1780 & \bf0.37 & 0.15 & 0 \\
\sc scholar + reg. & 1840 & 0.34 & 0.13 & 0.44 \\
\end{tabular}
\caption{Performance of various models on the Yahoo answers dataset with 250 topics and a 5,000-word vocabulary. SAGE did not finish in 72 hours so we omit it from this table. }
\label{tab:yahoo}
\end{table}

\begin{table*}
\centering
\begin{tabular}{r l }
NPMI & Topic \\
\hline
0.77 & turks armenian armenia turkish roads escape soviet muslim mountain soul \\
0.52 & escrow clipper encryption wiretap crypto keys secure chip nsa key \\
0.49 & jesus christ sin bible heaven christians church faith god doctrine \\
0.43 & fbi waco batf clinton children koresh compound atf went fire \\
0.41 & players teams player team season baseball game fans roger league \\
0.39 & guns gun weapons criminals criminal shooting police armed crime defend \\
0.37 & playoff rangers detroit cup wings playoffs montreal toronto minnesota games \\
0.36 & ftp images directory library available format archive graphics package macintosh \\
0.33 & user server faq archive users ftp unix applications mailing directory \\
0.32 & bike car cars riding ride engine rear bmw driving miles \\
0.32 & study percent sexual medicine gay studies april percentage treatment published \\
0.32 & israeli israel arab peace rights policy islamic civil adam citizens \\
0.30 & morality atheist moral belief existence christianity truth exist god objective \\
0.28 & space henry spencer international earth nasa orbit shuttle development vehicle \\
0.27 & bus motherboard mhz ram controller port drive card apple mac \\
0.25 & windows screen files button size program error mouse colors microsoft \\
0.24 & sale shipping offer brand condition sell printer monitor items asking \\
0.21 & driver drivers card video max advance vga thanks windows appreciated \\
0.19 & cleveland advance thanks reserve ohio looking nntp western host usa \\
0.04 & banks gordon univ keith soon pittsburgh michael computer article ryan \\
\end{tabular}
\caption{Topics from the unsupervised \textsc{scholar} on the 20 newsgroups dataset, and the corresponding internal coherence values. }
\label{tab:topics}
\end{table*}

\end{document}